\documentclass{article}
\usepackage{spconf,amsmath,amssymb,graphicx,booktabs,multirow}
\usepackage{flafter} %
\usepackage[hidelinks]{hyperref}
\usepackage{pifont}      %
\usepackage{xcolor}      %
\newcommand{\ResultSettings}{32}

\newcommand{\ResultWithoutMax}{36.8}

\newcommand{\ResultChineseMax}{36.8}

\newcommand{\ResultWhisperBaseB}{35.07}
\newcommand{\ResultMaxWhisperB}{22.16}

\newcommand{\LocalThinkingBReduction}{2.57}

\newcommand{\CARThinkingBReduction}{17.27}

\newcommand{\methodname}{Agentic-GER}

\title{Agentic-GER: Terminology Recovery\\in Long-Form Speech Using Global Context}
\name{Yanqiao Zhu$^{1,2,\dagger}$\thanks{$^{\dagger}$This work was done during an internship at Alibaba Token Foundry.} \qquad Wupeng Wang$^{3}$ \qquad Zhifu Gao$^{3}$ \qquad Xiangang Li$^{3}$  \qquad Xie Chen$^{1,2,*}$\thanks{* indicates the corresponding author}}
\address{$^{1}$X-LANCE Lab, Shanghai Jiao Tong University \\
$^{2}$Shanghai Innovation Institute \\
$^{3}$Alibaba Token Foundry}

\begin{document}
\maketitle

\begin{abstract}
Recent advances in speech language models have improved automatic speech recognition (ASR) for long-form audio. 
However, accurately and consistently transcribing domain-specific terminology remains challenging.
Motivated by the world knowledge and contextual capability of large language models (LLMs), we propose {\methodname}, an LLM-based agent for terminology correction in long-form speech.
The agent uses global context from the full transcript to identify suspicious terms and resolve ambiguous hypotheses.
It selectively re-transcribes the source speech to check candidate corrections, and uses accepted edits to guide subsequent decisions.
Experiments with four LLMs and two ASR systems on GigaSpeechBench show consistent terminology improvements in both Chinese and English, with and without thinking.
On Chinese speech, {\methodname} achieves up to a \ResultWithoutMax\% relative reduction in biased character error rate (B-CER) over the Whisper baseline.

\end{abstract}


\begin{figure}[!t]
  \centering
  \includegraphics[width=0.8\columnwidth,keepaspectratio]{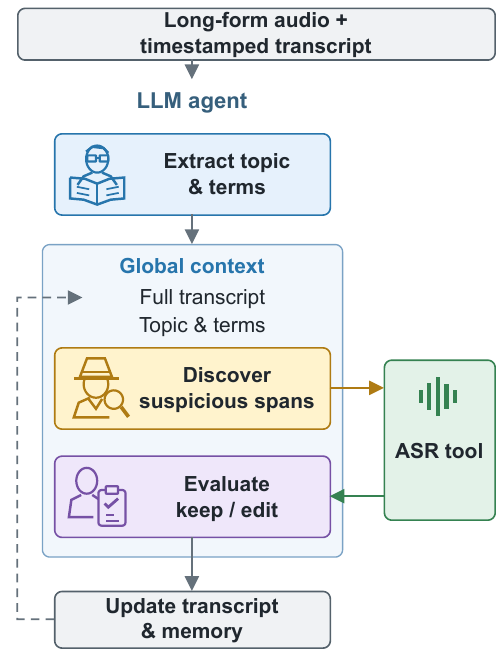}
  \caption{Overview of {\methodname}. Global context guides error discovery and correction. The agent re-transcribes selected segments to evaluate edits, updates the transcript, and records decisions for subsequent rounds.}
  \label{fig:pipeline}
\end{figure}
\vspace{-3pt}
\section{Introduction}
\label{sec:intro}
Automatic speech recognition (ASR) has made progress in recent years ~\cite{aed_whisper_radford,qwen3asr_shi,funasr_an}. Nevertheless, reliably transcribing long-form audio spanning minutes or hours remains challenging \cite{longspeech_bench}. A primary bottleneck is the accurate and consistent transcription of domain-specific terminology. In lengthy recordings, rare or specialized terms are susceptible to acoustic confusion and contextual dilution, leading to inconsistent hypotheses across different segments of the same audio.

Generative error correction (GER) uses the world knowledge and contextual capabilities of large language models (LLMs) to revise hypotheses without modifying the ASR model. Early approaches use textual evidence such as N-best lists~\cite{chen2023hyporadise,yang2023generative}; later work incorporates source speech~\cite{hu-etal-2024-listen}, retrieved entities, and contextual memory~\cite{ghosh-etal-2025-failing,li2026ontology}. 

Despite these advancements, applying such methods to hours-long audio requires bridging error discovery with correction.
This involves deciding which segments to re-transcribe and using context from the full transcript to resolve ambiguous ASR hypotheses.
LLM agents~\cite{yao2022react} can call tools to inspect selected audio segments while using the full transcript as context.

We view long-form correction as selective evidence acquisition guided by the full transcript. The transcript reveals recurring entities, related concepts, and cross-segment inconsistencies that indicate which parts of the source speech to inspect. This global context also constrains the candidate terms supplied by the LLM.

Based on this insight, we propose {\methodname}.\footnote{Code: \url{https://github.com/QwenAudio/FunResearch}}. As shown in Figure~\ref{fig:pipeline}, it scans the full transcript for suspicious spans, selectively re-transcribes the corresponding audio segments, and evaluates candidate edits using the resulting hypotheses and global context. Accepted edits update the transcript for subsequent rounds, allowing recovered terminology to inform later decisions.

We evaluate {\methodname} on GigaSpeechBench~\cite{GigaSpeechBench} across four LLMs, two languages, and two initial ASR systems, with and without thinking. We measure terminology recognition using biased character error rate (B-CER) for Chinese and biased word error rate (B-WER) for English, which focus on annotated terminology in the reference transcripts. {\methodname} consistently lowers these error rates in all \ResultSettings{} configurations, with larger gains in Chinese than in English. With Whisper, the relative B-CER reduction reaches \ResultChineseMax\% on Chinese speech.

On FunASR transcripts for Chinese speech, a comparison with local GER using Qwen3.8-27B supports the value of global context: {\methodname} improves both terminology and overall accuracy, whereas local GER trades small terminology gains for higher overall error rates.
Thinking consistently improves the 27B/31B correctors but provides limited or negative gains for Flash and Max.

\vspace{-3pt}

\section{Method}
\label{sec:method}

\begin{figure}[!t]
  \centering
  \includegraphics[width=0.95\columnwidth,keepaspectratio]{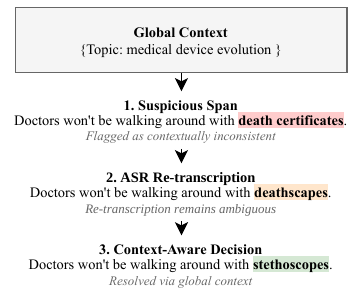}
  \caption{Resolving an ambiguous ASR hypothesis using global context. The agent flags ``death certificates'' in a discussion of medical device evolution. Re-transcription yields ``deathscapes.'' An earlier segment mentions a ``stethoscope.'' The agent corrects the term to ``stethoscopes,'' matching the reference. Excerpts are shortened for display.}
  \label{fig:case}
\end{figure}

{\methodname} takes audio $A$ and a timestamped transcript
$T^{(0)}=\{(s_i,e_i,x_i)\}_{i=1}^{N}$, where $x_i$ covers the interval
$[s_i,e_i]$. It uses the full transcript to find suspicious terms, then
checks them against the source speech through segment re-transcription
before editing the transcript.
Segment boundaries remain fixed throughout correction.

\subsection{Error Discovery via Global Context}
\label{sec:method:context}

The topic of the speech narrows the set of plausible terms. Motivated by contextual
prompting~\cite{suh24_interspeech,li2023prompting}, the agent first extracts the topic, domain, and terms from the initial
transcript. This summary remains fixed, supplying domain cues and observed term
forms.

Next, the agent uses this context to find recognition errors that may be
unremarkable within a single segment. It examines the summary, current
full transcript, and a working memory of earlier decisions, proposing
up to $k$ suspicious spans per round.
Each proposal specifies a segment, a short substring, and a reason for
inspection. Typical cues include inconsistent spellings of the same entity, phonetic confusions, and common words appearing in unlikely technical contexts. The proposed alternatives are evaluated in the next step.

\subsection{Re-transcription and Evaluation}
\label{sec:method:verification}

A term that fits the topic may differ from what was spoken.
For each proposed segment, the agent extracts $A[s_i:e_i]$ and
re-transcribes it to obtain a second hypothesis.

Finally, the agent decides whether to retain or revise each span. It
receives the current segment, its re-transcription, the candidate--evidence pairs, and the global context. The prompt asks the LLM
to assess pronunciation compatibility and support from the re-transcription
or consistent mentions elsewhere. When the re-transcription remains
ambiguous, global context can help resolve it.
The prompt requests a minimal edit, preserving unrelated words and
retaining the original if evidence is insufficient. The output contains
a keep/edit decision, the resulting segment, and a short explanation.
Figure~\ref{fig:case} illustrates how global context helps resolve an
ambiguous re-transcription.

\subsection{Iterative Correction}
\label{sec:method:execution}

Recovering one occurrence makes that term available when examining other
mentions. After all candidates in a round have been evaluated using the
same transcript, the agent applies the accepted changes. Working
memory records both edits and keep decisions, including the checked spans
and their explanations. The next round uses this memory and the updated 
transcript. The loop ends when no new candidates are found or the limit
on rounds or accepted edits is reached.

\vspace{-8pt}
\section{Experimental Setup}
\label{sec:setup}

We evaluate on GigaSpeechBench, using 524 Chinese
and 387 English audio files, totaling approximately 120 hours per language
across 12 domains. Initial
transcripts come from FunASR-Realtime~\cite{funasr_an} and
Whisper-Large-v3~\cite{aed_whisper_radford}. The four LLMs are
Qwen3.8-27B, Gemma4-31B, Qwen3.8-Flash, and Qwen3.8-Max, each evaluated
\emph{w/o Thinking} and \emph{w/ Thinking}. Thinking denotes
intermediate token generation before the task output. This yields 32 configurations.
Qwen3.8-27B and Gemma4-31B are dense models, whereas Flash and Max
are mixture-of-experts (MoE) models with 125B and 2.4T backbone
parameters, activating 6B and 95B parameters per token, respectively.
All configurations use Qwen3-ASR-1.7B~\cite{qwen3asr_shi}
for segment re-transcription, without task-specific fine-tuning.
Qwen3.8-27B, Gemma4-31B, and Qwen3-ASR-1.7B are served
via vLLM on a server with four NVIDIA H100
GPUs. Flash and Max are accessed through the Alibaba Cloud Bailian API.
We set $k=4$ for Chinese and $k=8$ for English, with at most 32
rounds and 48 accepted edits per audio file in all configurations.

Following GigaSpeechBench~\cite{GigaSpeechBench}, we report B-CER for
Chinese terminology and B-WER for English terminology.
Both are computed as $(S_b+D_b+I_b)/N_b\times100\%$, where
$S_b$, $D_b$, and $I_b$ count substitutions, deletions, and insertions
associated with annotated terminology under the benchmark's scoring
protocol, and $N_b$ counts the corresponding reference characters or
words. We follow the benchmark's normalization and annotations.
We also evaluate overall CER for Chinese and WER for English to assess
how terminology correction affects full-transcript accuracy.
Scores are macro-averaged over 12 domains.

\providecommand{\cmark}{\textcolor{green!60!black}{\ding{51}}}
\providecommand{\xmark}{\textcolor{red!70!black}{\ding{55}}}
\begin{table}[t]
\centering
\caption{Main terminology results without thinking. Error rates (\%, $\downarrow$); parentheses show relative error reductions (\%, $\uparrow$) from Baseline. Column minima are in bold.}
\label{tab:main}
\vspace{3pt}
\small
\setlength{\tabcolsep}{2pt}
\renewcommand{\arraystretch}{1.10}
\begin{tabular*}{\columnwidth}{@{\extracolsep{\fill}}lrrrr@{}}
\toprule
& \multicolumn{2}{c}{Chinese} & \multicolumn{2}{c}{English} \\
& \multicolumn{2}{c}{B-CER$\downarrow$} & \multicolumn{2}{c}{B-WER$\downarrow$} \\
\cmidrule(lr){2-3}\cmidrule(lr){4-5}
\phantom{zyq} & FunASR & Whisper & FunASR & Whisper \\
\midrule
Baseline & 10.91 & 35.07 & 12.94 & 14.67 \\
\midrule
\phantom{yq}+Qwen3.8-27B & \shortstack[r]{10.19\\(6.7)} & \shortstack[r]{28.88\\(17.7)} & \shortstack[r]{12.35\\(4.6)} & \shortstack[r]{14.08\\(4.0)} \\
\addlinespace[3pt]
\phantom{yq}+Gemma4-31B & \shortstack[r]{9.88\\(9.5)} & \shortstack[r]{25.09\\(28.5)} & \shortstack[r]{12.05\\(6.9)} & \shortstack[r]{13.71\\(6.5)} \\
\addlinespace[3pt]
\phantom{yq}+Qwen3.8-Flash & \shortstack[r]{9.64\\(11.7)} & \shortstack[r]{26.61\\(24.1)} & \shortstack[r]{11.91\\(8.0)} & \shortstack[r]{13.72\\(6.5)} \\
\addlinespace[3pt]
\phantom{yq}+Qwen3.8-Max & \shortstack[r]{\textbf{8.96}\\(17.9)} & \shortstack[r]{\textbf{22.16}\\(36.8)} & \shortstack[r]{\textbf{11.74}\\(9.3)} & \shortstack[r]{\textbf{13.64}\\(7.0)} \\
\bottomrule
\end{tabular*}
\end{table}

\vspace{-5pt}
\section{Results and Analysis}
\label{sec:results}

\subsection{Recovering Domain Terminology}
\label{sec:results:main}

Table~\ref{tab:main} reports results without
thinking. {\methodname} improves B-CER/B-WER for every evaluated
LLM under both ASR systems. The larger Flash and Max models generally
yield greater terminology gains than the 27B/31B models, with Max
performing best in all four ASR/language combinations.
On Chinese speech, Max reduces B-CER for
Whisper from \ResultWhisperBaseB\% to \ResultMaxWhisperB\%, a
\ResultWithoutMax\% relative reduction. The gains also extend to
FunASR, whose baseline terminology error rates are lower. This shows
that terminology errors remain correctable even with a stronger ASR
baseline. Relative gains are larger for Chinese than for English under
both ASR systems, motivating a closer look at the errors being corrected.

We also evaluate overall transcription accuracy across both thinking
modes. For Whisper on Chinese speech, overall CER decreases by
8.7--18.3\% relative to the baseline. For the other configurations, relative changes in overall CER/WER range from a
3.4\% decrease to a 0.5\% increase.

\subsection{Error-Type Analysis}
\label{sec:results:error_types}

To investigate the larger gains on Chinese speech, we compare
substitution and deletion errors across the two languages.
Table~\ref{tab:error_types} reports baseline error proportions and
relative reductions in error counts for Qwen3.8-27B on Whisper
transcripts without thinking. 
We pool terminology alignment errors
across the evaluation data, using characters for Chinese and words
for English.

Substitution errors show larger relative reductions than deletions in both languages. For Whisper on Chinese
speech, substitution errors decrease by 21.3\%, compared with 3.0\%
for deletions; on English speech, the corresponding reductions are
6.4\% and 0.4\%. FunASR shows the same pattern, including a 5.4\%
increase in deletions on Chinese speech. Substitution reductions
contribute most to the gains in all 32 configurations.

\providecommand{\cmark}{\textcolor{green!60!black}{\ding{51}}}
\providecommand{\xmark}{\textcolor{red!70!black}{\ding{55}}}
\begin{table}[!htbp]
\centering
\caption{Terminology error analysis for Qwen3.8-27B on Whisper
without thinking. Proportion (\%) is each error type's percentage
of all baseline terminology errors, including insertions.
Reduction (\%,$\uparrow$) is the relative decrease in its error count.}
\label{tab:error_types}
\vspace{3pt}
\small
\setlength{\tabcolsep}{4pt}
\renewcommand{\arraystretch}{1.05}
\begin{tabular}{@{}llrr@{}}
\toprule
Language & Error type & Proportion & Reduction$\uparrow$ \\
\midrule
\multirow{2}{*}{Chinese} & Substitution & 85.0 & 21.3 \\
 & Deletion & 14.7 & 3.0 \\
\midrule
\multirow{2}{*}{English} & Substitution & 56.0 & 6.4 \\
 & Deletion & 43.1 & 0.4 \\
\bottomrule
\end{tabular}
\end{table}

This pattern is consistent with how the agent discovers errors.
A substitution can leave an inconsistent term or an implausible phrase to inspect. An omitted term may
leave no textual cue, making it harder to identify through
transcript inspection. In the Whisper baseline, deletions account for
43.1\% of terminology errors in English, compared with 14.7\% in
Chinese; FunASR shows a similar difference. We hypothesize that this
higher proportion of deletions contributes to the smaller gains on
English transcripts.

\subsection{The Value of Global Context}
\label{sec:results:ablation}

To assess the value of global context, we compare {\methodname} with
local GER using Qwen3.8-27B on FunASR transcripts for Chinese speech.
Local GER corrects each segment in a single step from its original
text and the Qwen3-ASR hypothesis, without global context or
iterative updates. Table~\ref{tab:ablation} reports both thinking modes.

\begin{table}[!htbp]
\centering
\caption{Local GER versus {\methodname}. Error rates (\%, $\downarrow$); best values in bold. 
``Thinking'' indicates whether intermediate token generation is enabled (\cmark) or disabled (\xmark).}
\label{tab:ablation}
\vspace{3pt}
\setlength{\tabcolsep}{1pt}
\renewcommand{\arraystretch}{1}
\begin{tabular*}{\columnwidth}{@{\extracolsep{\fill}}lcrr@{}}
\toprule
\phantom{zyq} & Thinking & B-CER$\downarrow$ & CER$\downarrow$ \\
\midrule
Baseline & --- & 10.91 & 3.12 \\
\midrule
\multirow{2}{*}{\phantom{yq}+Local GER} 
    & \xmark & 10.80 & 3.37 \\
    & \cmark & 10.63 & 3.42 \\
\addlinespace[3pt]
\multirow{2}{*}{\phantom{yq}+{\methodname}} 
    & \xmark & 10.19 & 3.11 \\
    & \cmark & \textbf{9.03} & \textbf{3.01} \\
\bottomrule
\end{tabular*}
\end{table}

Local GER yields small terminology gains but increases overall CER,
whereas {\methodname} reduces both metrics. With thinking, the relative
B-CER reductions are \LocalThinkingBReduction\% for local GER and
\CARThinkingBReduction\% for {\methodname}. A second ASR hypothesis
can help recover a term, but may contain errors, as
Figure~\ref{fig:case} illustrates. Global context provides
constraints for choosing a replacement. The contrast between the two
workflows supports using the full transcript to select suspicious spans
and judge replacements across iterations, improving terminology and overall accuracy.

\subsection{Thinking for Terminology Recovery}
\label{sec:results:thinking}
\providecommand{\cmark}{\textcolor{green!60!black}{\ding{51}}}
\providecommand{\xmark}{\textcolor{red!70!black}{\ding{55}}}
\begin{table}[t]
\centering
\caption{Terminology error rates (\%, $\downarrow$) with thinking. $\Delta$ Thinking reports the change in relative error reduction compared with Table~\ref{tab:main} (percentage points, $\uparrow$). Positive values indicate improvement; negative values (bold) indicate degradation.}
\label{tab:thinking}
\vspace{3pt}
\small
\setlength{\tabcolsep}{2pt}
\renewcommand{\arraystretch}{1.10}
\begin{tabular*}{\columnwidth}{@{\extracolsep{\fill}}lrrrr@{}}
\toprule
& \multicolumn{2}{c}{Chinese} & \multicolumn{2}{c}{English} \\
& \multicolumn{2}{c}{B-CER$\downarrow$} & \multicolumn{2}{c}{B-WER$\downarrow$} \\
\cmidrule(lr){2-3}\cmidrule(lr){4-5}
\phantom{zyq} & FunASR & Whisper & FunASR & Whisper \\
\midrule
\phantom{yq}+Qwen3.8-27B & 9.03 & 25.89 & 11.96 & 13.87 \\
$\Delta$ Thinking (pp)$\uparrow$ & $+10.6$ & $+8.5$ & $+3.0$ & $+1.5$ \\
\addlinespace[3pt]
\phantom{yq}+Gemma4-31B & 9.32 & 23.20 & 11.79 & 13.60 \\
$\Delta$ Thinking (pp)$\uparrow$ & $+5.1$ & $+5.4$ & $+2.0$ & $+0.8$ \\
\addlinespace[3pt]
\phantom{yq}+Qwen3.8-Flash & 9.61 & 26.75 & 11.78 & 13.63 \\
$\Delta$ Thinking (pp)$\uparrow$ & $+0.3$ & $\boldsymbol{-0.4}$ & $+0.9$ & $+0.6$ \\
\addlinespace[3pt]
\phantom{yq}+Qwen3.8-Max & 9.45 & 25.54 & 11.79 & 13.62 \\
$\Delta$ Thinking (pp)$\uparrow$ & $\boldsymbol{-4.5}$ & $\boldsymbol{-9.7}$ & $\boldsymbol{-0.4}$ & $+0.1$ \\
\bottomrule
\end{tabular*}
\end{table}

We next examine whether thinking improves correction over the results
in Table~\ref{tab:main}. Table~\ref{tab:thinking} reports error rates
with thinking and $\Delta$ Thinking, the change in relative error
reduction in percentage points. Thinking improves Qwen3.8-27B and
Gemma4-31B in all four ASR/language combinations. Flash shows small
gains in three comparisons, whereas Max regresses in three. For Whisper
on Chinese speech, thinking increases the relative B-CER reduction by
8.5 percentage points for Qwen3.8-27B but decreases it by 9.7 points
for Max.

We hypothesize that this pattern reflects a tradeoff between eliciting
relevant terminology and introducing competing alternatives.
Kambhampati et al.~\cite{kambhampati2026intermediate} interpret
intermediate tokens as learned prompt augmentations. For Qwen3.8-27B
and Gemma4-31B, this augmentation may help bring relevant terms into
context and connect them to the speech. Max already performs best
without thinking, suggesting less need for additional elicitation.
Further generation may introduce alternatives to a suitable correction. This interpretation is consistent with the
association between incorrect answers and increased backtracking
reported by
Hassid et al.~\cite{hassid2025overthink} on reasoning benchmarks.

Beyond its effect on accuracy, thinking also increases correction time,
with a much larger rise in output-token usage than in input-token usage.
For example, on Whisper transcripts for Chinese speech, Gemma4-31B uses
161 million input tokens and 2.6 million output tokens without thinking,
with a correction RTF of 0.42. With thinking, input usage rises to
198 million tokens and output usage to 38.3 million, while RTF reaches 5.71.

\vspace{-5pt}
\section{Conclusion}
\label{sec:conclusion}

We presented {\methodname}, which uses global context to guide terminology
correction in long-form speech. The full transcript helps the agent locate
suspicious spans and judge candidate replacements together with ASR
re-transcriptions. Accepted edits update the transcript, allowing recovered
terms to inform subsequent decisions. Experiments on GigaSpeechBench show
terminology improvements across all 32 configurations. Substitution
reductions contribute most to these gains, while deletions show smaller
improvements. Thinking consistently benefits the 27B/31B models but yields
limited or negative gains for Flash and Max. 

A promising direction is to expand the agent's tool set. Web search could
provide domain knowledge about unfamiliar terms, while end-to-end
multimodal models could assess candidate corrections directly from speech
and transcript context. Future work will also investigate the recovery of
omitted terminology and evaluate the framework on additional datasets.

\bibliographystyle{IEEEbib}
\bibliography{references}
\end{document}